%% file: main.tex
\documentclass{article}

\usepackage[preprint]{neurips_2026}
\setcitestyle{numbers,square,comma,sort&compress}

\usepackage[utf8]{inputenc}
\usepackage[T1]{fontenc}
\usepackage{hyperref}
\usepackage{url}
\usepackage{booktabs}
\usepackage{amsfonts}
\usepackage{amsmath}
\usepackage{amssymb}
\usepackage{amsthm}

\usepackage{nicefrac}
\usepackage{microtype}
\usepackage[table]{xcolor}
\usepackage{graphicx}
\usepackage{twemojis}
\usepackage{fontawesome5}
\usepackage{caption}
\usepackage{subcaption}
\usepackage{multirow}
\usepackage{array}
\usepackage{wrapfig}
\usepackage{pgfplots}
\usepackage{tcolorbox}
\pgfplotsset{compat=1.18}

\definecolor{titlepanelbg}{HTML}{F2F6FA}

\newcommand{\ig}{\Delta H}                      
\newcommand{\igt}{\ig_t}                        
\newcommand{\pith}{\pi}                         
\newcommand{\piS}{\pith}                        
\newcommand{\piT}{\pith}                        
\newcommand{\pinfo}{z}                          
\newcommand{\sdpo}{SD}
\newcommand{\sdpoig}{SD+mask} 
\newcommand{\drgrpo}{Dr.\,GRPO}
\newcommand{\grpo}{GRPO}
\newcommand{\dapo}{DAPO}
\newcommand{\rmbench}{\textsc{RM-Bench}}
\newcommand{\rewardbench}{\textsc{RewardBench\,v2}}
\newcommand{\helpsteer}{\textsc{HelpSteer3-Preference}}

\newcommand{\verdict}[1]{{\setlength{\fboxsep}{1.5pt}\colorbox{gray!15}{#1}}}
\newcommand{\best}[1]{\cellcolor{gray!30}\textbf{#1}}
\newcommand{\snd}[1]{\cellcolor{gray!15}#1}

\title{Training LLM Judges from Language Feedback via Position-Selective Self-Distillation}

\author{%
  \bfseries
  Ilgee Hong\textsuperscript{1,\dag}\quad
  Changlong Yu\textsuperscript{2}\quad
  Zhenghao Xu\textsuperscript{1,\dag}\quad
  Xin Liu\textsuperscript{2} \\[2pt]
  \bfseries
  Yuwei Zhang\textsuperscript{3,\dag}\quad
  Qin Lu\textsuperscript{2}\quad
  Bing Yin\textsuperscript{2}\quad
  Tuo Zhao\textsuperscript{2} \\[8pt]
  \normalfont
  \textsuperscript{1}Georgia Institute of Technology\quad
  \textsuperscript{2}Amazon\quad
  \textsuperscript{3}UC San Diego
}

\newcommand{\blfootnote}[1]{%
  \begingroup
  \renewcommand{\thefootnote}{}%
  \footnotetext{#1}%
  \endgroup
  \addtocounter{footnote}{-1}%
}

\makeatletter
\renewcommand{\@maketitle}{%
  \vskip 0.1in
  \noindent
  \begingroup
  \begin{tcolorbox}[
    colback=titlepanelbg,
    colframe=titlepanelbg,
    boxrule=0pt,
    arc=8pt,
    outer arc=8pt,
    left=14pt,
    right=14pt,
    top=14pt,
    bottom=14pt,
    boxsep=0pt,
    width=\textwidth
  ]
    \begin{minipage}{\linewidth}
      \centering
      {\LARGE\bfseries \@title\par}
      \vspace{2\baselineskip}
      \if@anonymous
        \begin{tabular}[t]{c}\bfseries
          Anonymous Author(s) \\
          Affiliation \\
          Address \\
          \texttt{email}
        \end{tabular}%
      \else
        \begin{tabular}[t]{c}\bfseries\@author\end{tabular}%
      \fi
      \par\vspace{\baselineskip}
      \href{https://huggingface.co/opsd-genrm}{\texttwemoji{hugs}\,\textbf{Model}}
      \quad
      \href{https://github.com/horizon-llm/OPSD-GenRM}{\faGithub\,\textbf{Code}}
      \renewenvironment{abstract}%
        {\par\vspace{\baselineskip}%
         \begin{list}{}{%
           \setlength{\leftmargin}{8pt}%
           \setlength{\rightmargin}{8pt}%
           \setlength{\listparindent}{0pt}%
           \setlength{\itemindent}{0pt}%
           \setlength{\parsep}{0pt}%
         }\item\relax}%
        {\par\end{list}\vskip 1ex}%
      \input{sections/abstract.tex}
    \end{minipage}%
  \end{tcolorbox}%
  \endgroup
  \vskip 0.12in
}
\makeatother

\begin{document}

\maketitle
\blfootnote{\textsuperscript{\dag}Work done during internship at Amazon.}

\noindent
\begin{minipage}{\linewidth}
  \centering
  \begin{tikzpicture}
    \begin{axis}[
      width=0.48\linewidth, height=3.8cm,
      xlabel={Training step}, ylabel={\rmbench{} Acc. (\%)},
      legend to name=sharedlegend,
      legend columns=4, legend style={font=\small, draw=none},
      grid=major, grid style={gray!20},
      title={Qwen3-4B-Instruct}, title style={font=\small},
      xmin=0, xmax=220, ymin=81, ymax=87,
    ]
      \addplot[thick, blue!60!black, densely dashed] coordinates {(0,81.70)(20,82.23)(40,82.37)(60,82.55)(80,82.84)(100,82.83)(120,82.78)(140,82.84)(160,82.90)(180,83.03)(200,83.36)(220,83.47)};
      \addplot[thick, blue] coordinates {(0,81.70)(20,82.79)(40,83.26)(60,83.58)(80,83.60)(100,83.80)(120,84.07)(140,84.17)(160,84.30)(180,84.33)(200,84.35)(220,84.28)};
      \addplot[thick, color={rgb,255:red,213;green,94;blue,0}, dashed] coordinates {(0,81.70)(20,81.95)(40,81.99)(60,82.25)(80,82.49)(100,82.55)(120,82.32)(140,82.19)(160,82.24)(180,82.24)(200,82.38)(220,82.26)};
      \addplot[thick, black!60, dotted] coordinates {(0,85.99)(220,85.99)};
      \legend{\sdpo{}, SD+mask 70\%, \drgrpo{}, DeepSeek-R1}
    \end{axis}
  \end{tikzpicture}
  \hfill
  \begin{tikzpicture}
    \begin{axis}[
      width=0.48\linewidth, height=3.8cm,
      xlabel={Training step}, ylabel={\rmbench{} Acc. (\%)},
      grid=major, grid style={gray!20},
      title={Qwen3-30B-A3B-Instruct}, title style={font=\small},
      xmin=0, xmax=220, ymin=83, ymax=89,
    ]
      \addplot[thick, blue!60!black, densely dashed] coordinates {(0,84.10)(20,84.34)(40,84.41)(60,84.41)(80,84.59)(100,84.74)(120,84.66)(140,84.84)(160,84.89)(180,84.86)(200,84.91)(220,85.15)};
      \addplot[thick, blue] coordinates {(0,84.10)(20,85.45)(40,85.72)(60,86.06)(80,86.41)(100,86.50)(120,86.67)(140,86.79)(160,86.87)(180,86.90)(200,87.07)(220,87.14)};
      \addplot[thick, color={rgb,255:red,213;green,94;blue,0}, dashed] coordinates {(0,84.10)(20,84.36)(40,84.35)(60,84.03)(80,84.00)(100,84.43)(120,84.86)(140,85.09)(160,85.48)(180,85.68)(200,85.82)(220,85.80)};
      \addplot[thick, black!60, dotted] coordinates {(0,85.99)(220,85.99)};
    \end{axis}
  \end{tikzpicture}\\[4pt]
  \ref{sharedlegend}
  
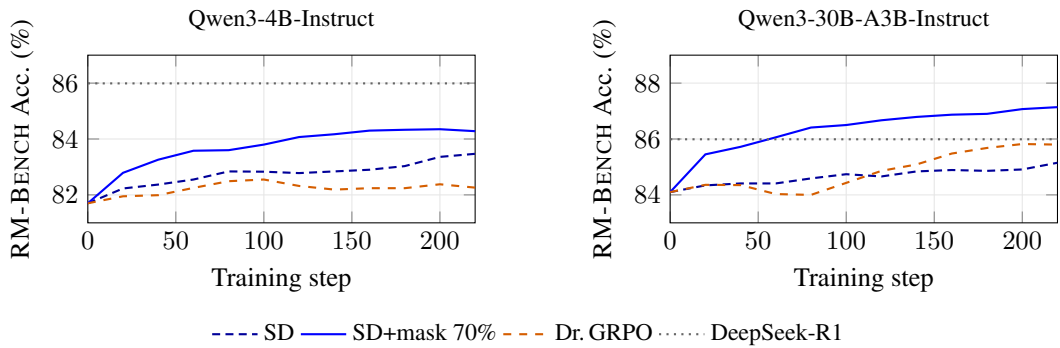
\captionof{figure}{Out-of-distribution \rmbench{} accuracy during training. Curves show the debiased EMA ($\beta = 0.6$) of accuracy evaluated every 20 training steps. All methods start from a judge prompt template tuned for best base-model accuracy. \sdpoig{} outperforms naive \sdpo{} and \drgrpo{}.}
  \label{fig:headline}
\end{minipage}

\input{sections/intro.tex}
\input{sections/related.tex}
\input{sections/method.tex}

\input{sections/experiments.tex}

\input{sections/conclusion.tex}
\input{sections/limitations.tex}

\bibliographystyle{plainnat}
\bibliography{references}

\newpage
\appendix
\input{sections/appendix.tex}

\end{document}

%% file: sections/abstract.tex
\begin{abstract}
We study training LLM judges from natural language feedback, especially for subjective tasks where the verdict depends strongly on which evaluation criteria the judge invokes and how it weighs them.
The dominant approach, outcome-supervised RL (e.g., \grpo{}), credits every token in the rollout with a single scalar determined only by the accuracy of the final verdict, providing no separate credit at the criterion-choice tokens and ignoring the rich language feedback (e.g., preference rationales) that naturally accompanies preference labels.
Self-Distillation (\sdpo{}) is one natural way to use this language feedback: the same model, conditioned on this feedback, acts as a teacher providing dense, position-level supervision. However, not all positions carry equally useful signal.
Using the per-position entropy shift between teacher and student, we identify two regimes: \emph{context sharpening}, where the teacher concentrates probability on a particular feedback-aligned criterion expression, and \emph{context spreading}, where the teacher distributes probability across multiple feedback-aligned alternatives. We interpret these patterns as follows: sharpening encourages memorization of a particular criterion expression, whereas spreading promotes semantic understanding by preserving these alternatives. Motivated by this asymmetry, we introduce position masking based on the entropy shift that retains the lower tail of the entropy-shift distribution.
Experiments show that masking higher-entropy-shift positions improves out-of-distribution generalization over naive \sdpo{}. The resulting self-distilled judges outperform judges trained with outcome-supervised RL by 2--9 percentage points on the evaluated subjective subcategories, while remaining competitive on objective ones.
\end{abstract}

%% file: sections/intro.tex
\section{Introduction}
\label{sec:intro}

Training LLMs to \emph{judge} responses, both as standalone evaluators and as reward models for downstream training, is a core building block of modern post-training.
A judge's verdict varies along two distinct axes: 1)~\emph{criterion choice}, i.e., which evaluation criteria the model invokes and how it weighs them, and 2)~\emph{application rigor}, i.e., how rigorously those criteria are applied to evaluate responses.
Judgment tasks fall into two regimes by which of these two axes decides the verdict: \emph{objective} tasks, where the decisive criterion is clear-cut (e.g., correctness), so the verdict depends on application rigor (e.g., math, coding), and \emph{subjective} tasks, where the decisive criterion is subtle and multifaceted (e.g., what counts as helpful for this chat), so the choice and weighting of criteria can drive the verdict.
Outcome-supervised RL (e.g., \grpo{}~\citep{shao2024deepseekmath}, \drgrpo{}~\citep{liu2025drgrpo}, \dapo{}~\citep{yu2025dapo}) with a single verdict-correctness reward is the dominant approach to train LLM judges~\citep{whitehouse2025j1,hong2025thinkrm,guo2025rrm,chen2026rmr1,wang2026outcome,xu2026alternating}, and works well on objective tasks by sharpening reasoning over a clear-cut criterion.

For subjective tasks, however, outcome-supervised RL provides limited explicit guidance on criterion choice.
It credits every token in the rollout with a single scalar determined only by the accuracy of the final verdict; it shapes criterion selection and weighting indirectly, through whether a given choice produces a correct verdict, without explicitly distinguishing the contributions of criterion choice and criterion application.
Even with high application rigor, a model that applies the \emph{wrong} criterion (or weighs competing criteria poorly) still produces a wrong verdict.
Compounding the problem, outcome-supervised RL is known to drive policy entropy downward over training~\citep{cui2025entropy,yu2025dapo}, which further narrows the pool of criteria the judge explores.

Many preference datasets naturally contain per-example natural language feedback alongside the preference label, since obtaining the label typically requires annotators to articulate why one response is preferred over the other: preference rationales that explicitly name the decisive criterion~\citep{wang2025helpsteer3,liu2025openrubrics}, a signal that the outcome-only objective leaves unused.
Self-Distillation (\sdpo{})~\citep{huebotter2026sdpo,zhao2026selfdistilledreasoner} is one natural way to use this language feedback: the same model, conditioned on this feedback, acts as a teacher and provides dense distributional guidance at every position of the student's rollouts. Unlike outcome-supervised RL, this dense per-position supervision now provides separate credit at the criterion-choice tokens, since the teacher's distributional signal at those positions is directly shaped by the language feedback that names the decisive criterion.

Not every position in the distillation loss carries equally useful signal.
To characterize this heterogeneity, we define the per-position \emph{entropy shift} as the entropy reduction in the teacher's next-token distribution relative to the student's, induced by conditioning the teacher on the language feedback, and use it to identify two regimes.
At positions with a large positive entropy shift (\emph{context sharpening}), the teacher concentrates probability on a particular feedback-aligned criterion expression.
At positions with a large negative entropy shift (\emph{context spreading}), the teacher distributes probability across multiple feedback-aligned alternatives.
We interpret these patterns as follows: sharpening encourages memorization of a particular criterion expression, whereas spreading promotes semantic understanding by preserving these alternatives.
Figures~\ref{tab:sharpening-examples},~\ref{tab:spreading-examples},~\ref{fig:sharpening-appx}, and~\ref{fig:spreading-appx} illustrate concrete examples of these two regimes.
Motivated by this asymmetry, we introduce \emph{position masking based on entropy shift}: retain the lower tail of each generation's entropy-shift distribution for the distillation loss, favoring context-spreading positions while removing the highest-shift positions.

Experiments show that self-distilled judges outperform judges trained with outcome-supervised RL (\drgrpo{}) by $2$--$9$ percentage points on the evaluated subjective subcategories, while \drgrpo{} remains competitive on objective subcategories where application rigor matters most. Further, masking higher-entropy-shift positions improves out-of-distribution (OOD) generalization over naive \sdpo{}.
As Figure~\ref{fig:headline} shows, \sdpo{}+mask leads both naive \sdpo{} and \drgrpo{} on \rmbench{}~\citep{liu2025rmbench} throughout training for Qwen3-4B-Instruct and Qwen3-30B-A3B-Instruct~\citep{yang2025qwen3}, surpassing strong reasoning judges including DeepSeek-R1~\citep{deepseekai2025r1} and Claude-Sonnet-4~\citep{anthropic2025claude4} at the 30B scale.

%% file: sections/related.tex
\section{Related Work}
\label{sec:related}

\paragraph{Learning from natural language feedback.}
Prior work uses natural language feedback primarily through refinement or critique pipelines. At inference time, models are prompted to iteratively revise outputs from self- or external-model-generated language feedback~\citep{madaan2023selfrefine,wadhwa2024learning,lee2025feedbackdescent}. At training time, models are fine-tuned on refinements that incorporate language feedback~\citep{scheurer2023learning}, on self-critiques and revisions generated from natural language principles~\citep{bai2022constitutional}, while other methods augment GRPO with additional refinement rollouts conditioned on a critique of the initial response~\citep{zhang2025critique}. Text2Grad~\citep{wang2025text2grad} instead aligns critique phrases with response spans and converts these alignments into per-span differentiable reward signals that drive gradient updates on the offending tokens. A more recent line of on-policy \sdpo{} uses the same model, additionally conditioned on language feedback unavailable to the student at inference, as the teacher. This framework incorporates diverse types of language feedback, including reference solutions, environmental feedback, successful rollouts, expert demonstrations, and dynamically summarized skills~\citep{zhao2026selfdistilledreasoner,huebotter2026sdpo,shenfeld2026sdft,wang2026skillsd}. Our work uses one- or two-sentence annotator rationales as language feedback for judge training. These rationales expose the evaluation criteria behind each preference label, which is especially useful for subjective tasks where generalization depends on selecting and weighting the right criteria.

\paragraph{Token selection methods for post-training.}
Recent work has begun to replace uniform token-level supervision with selective updates during LLM post-training. In supervised fine-tuning and preference optimization, several methods filter or reweight tokens based on influence-based quality, counterfactual importance, per-token KL, or preference-derived importance scores~\citep{pang2025token,ruan2025critical,zeng2024tdpo,liu2025tisdpo,yang2026tidpo}. In RLVR, high-entropy token selection identifies a small set of uncertain ``forking'' tokens that dominate policy-gradient learning, while polarity--entropy decomposition and gradient-magnitude selection further refine token-level credit assignment~\citep{wang2025minority,he2026polarity,lv2026gmts}. Closest to our setting, on-policy distillation methods select or reweight token losses using teacher entropy, student entropy and teacher--student divergence, log-probability gaps with LLM-judged relevance, training-trajectory dynamics, position-based teacher reliability, or asymmetric updates in non-positive-advantage regions~\citep{feng2026rmsd,jin2026entropyaware,xu2026tip,shen2026t3s,liu2026pwopsd,jia2026aopd}. In contrast, our method studies position selection in full-logit on-policy \sdpo{} for judge training via the entropy shift between the student and feedback-conditioned teacher distributions.

%% file: sections/method.tex
\section{Method}
\label{sec:method}

\subsection{Preliminary: Outcome-Supervised RL for LLM Judges}
\label{sec:rlvr-prelim}

A pairwise LLM judge is trained on examples of the form $(x, y_A, y_B, c^\star)$: a prompt $x$, two candidate responses $y_A, y_B$, and a gold preference label $c^\star \in \mathcal{C}$ where $\mathcal{C}$ is a finite set of possible verdicts.
The verdict set $\mathcal{C}$ could be simply binary (\verdict{$y_A \succ y_B$} or \verdict{$y_A \prec y_B$}) or multiclass~\citep{hong2025thinkrm,wang2026outcome}, for example including \verdict{\textsc{tie}}, \verdict{\textsc{unknown}/\textsc{unclear}}, or a multi-way ordinal preference such as \verdict{$y_A \succ\succ y_B$}, \verdict{$y_A \succ y_B$}, \verdict{$y_A \sim y_B$}. In this paper, we consider the binary setting with $\mathcal{C} = \{A, B\}$.
Given $(x, y_A, y_B)$, the judge generates a token sequence $\tau = (a_1, a_2, \ldots, a_T)$ one token at a time, $a_t \sim \pi(\cdot \mid s_t)$, where $s_t = (x, y_A, y_B, a_1, \ldots, a_{t-1})$ is the prefix at position $t$.
The sequence consists of an intermediate trace (criterion choice and application) and a final verdict $c \equiv a_T \in \mathcal{C}$.

The dominant training algorithm is outcome-supervised RL.
Each rollout is scored by whether its final verdict matches the gold preference label:
\begin{equation}
  R(\tau) \;=\; \mathbf{1}\!\left[c = c^\star\right].
  \label{eq:rlvr-reward}
\end{equation}
The judge is then optimized against this reward using GRPO-style policy-optimization algorithms~\citep{shao2024deepseekmath,liu2025drgrpo,yu2025dapo}. Even when a training example contains language feedback $\pinfo$, such as a preference rationale that articulates why one response is preferred over the other and identifies the decisive criterion, this outcome-only reward leaves $\pinfo$ unused. It also assigns the same outcome-based credit to every token position, with no direct supervision where the judge chooses and weighs evaluation criteria.

\subsection{Language Feedback Self-Distillation for LLM Judges}
\label{sec:sdpo-preliminaries}

To use the language feedback left unused by the outcome-only objective, we apply self-distillation (\sdpo{}) to provide dense, position-specific supervision for the judge's intermediate trace.

Each training example carries a piece of language feedback $\pinfo$ underlying its preference label $c^\star$. We use a single model $\pith$ in two roles: as the \emph{student}, conditioned on the prompt alone, $\pith(\cdot \mid s_t)$; and as the \emph{teacher}, which additionally conditions on $\pinfo$, $\pith(\cdot \mid s_t, \pinfo)$.
The context is \emph{privileged} in the sense that the student is never given $\pinfo$, either during training or at deployment.
We adopt the on-policy reverse-KL \sdpo{} objective from \citep{huebotter2026sdpo, zhao2026selfdistilledreasoner} as the underlying loss. Over a training set $\mathcal{D}$ of judge examples $(x, y_A, y_B, \pinfo)$ with on-policy rollouts $\tau = (a_1, \ldots, a_T) \sim \piS$ drawn per example,
\begin{equation}
  \mathcal{L}_{\text{SD}}(\pith) =
  \mathbb{E}_{(x,y_A,y_B,\pinfo)\sim\mathcal{D},\;\tau\sim\pith}
  \left[
    \frac{1}{T}\sum_{t=1}^{T}
    \mathrm{KL}\!\left(
      \pith(\cdot\mid s_t)
      \;\Big\|\;
      \mathrm{sg}\!\left[\pith(\cdot\mid s_t,\pinfo)\right]
    \right)
  \right],
  \label{eq:sdpo-loss}
\end{equation}
where $\mathrm{sg}[\cdot]$ is stop-gradient and the inner KL is a full-vocabulary sum at each position. Unlike a verdict-level reward, this objective transfers the feedback-conditioned teacher distribution at every token position, including positions where the judge selects evaluation criteria.

\begin{figure}[!t]
  \centering
  \begin{minipage}{0.85\linewidth}
    \begin{subfigure}[t]{0.46\linewidth}
      \includegraphics[width=\linewidth]{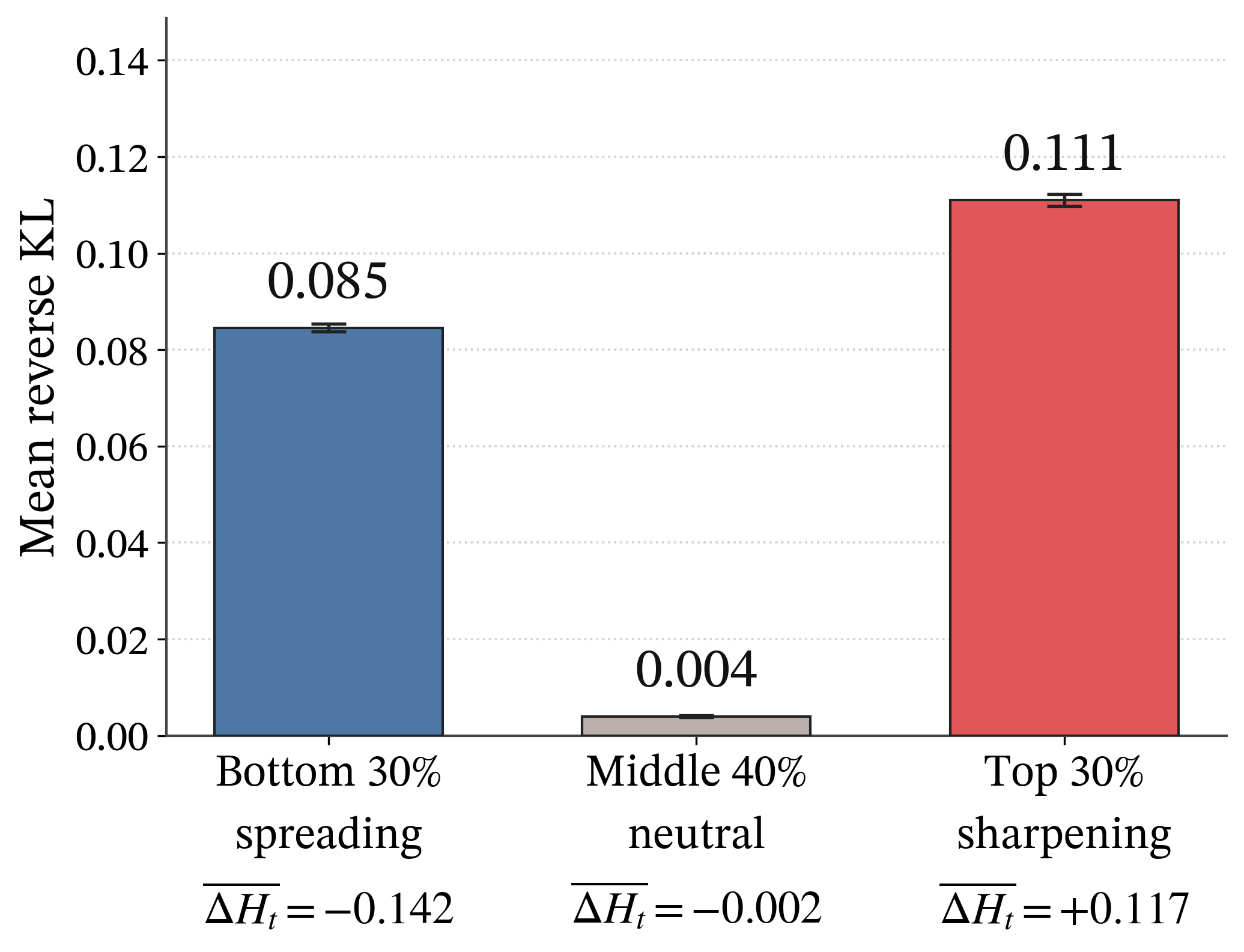}
      \caption{Mean reverse KL by $\igt$ tier.}
      \label{fig:revkl-by-ig-tier-a}
    \end{subfigure}\hfill
    \begin{subfigure}[t]{0.52\linewidth}
      \includegraphics[width=\linewidth]{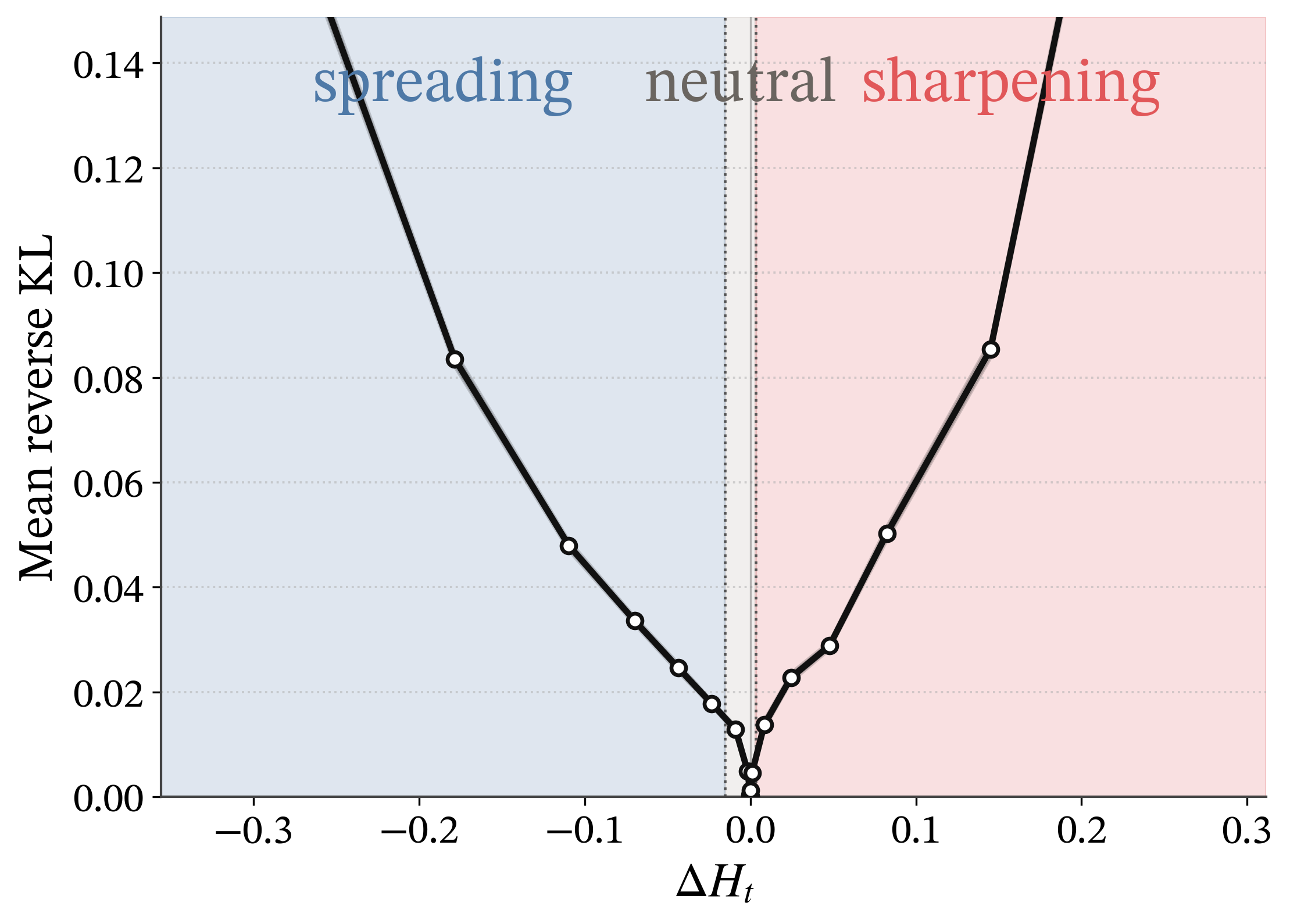}
      \caption{Mean reverse KL as a function of $\igt$.}
      \label{fig:revkl-by-ig-tier-b}
    \end{subfigure}
  \end{minipage}
  \caption{Both $\igt$ tails account for most of the distillation loss. \textbf{(a)} Three-way per-sequence percentile partition (top-30\% $\igt$ / middle 40\% / bottom-30\%): both tails carry $\sim$20--28$\times$ the per-position KL of the neutral middle. \textbf{(b)} Binned mean reverse KL along $\igt$ using 20 equal-count bins (95\% x-range shown).}
  \label{fig:revkl-by-ig-tier}
\end{figure}

\subsection{Analysis of Per-Position Self-Distillation Signal}
\label{sec:theory}

Eq.~\ref{eq:sdpo-loss} treats every response position as an equally valid imitation target, but the language feedback $\pinfo$ does not affect the teacher uniformly across positions.
We characterize this heterogeneity through the per-position entropy shift, then examine its connection to criterion choice.
As discussed above, language feedback identifies the decisive criteria behind a preference label, allowing self-distillation to provide direct supervision at criterion-choice tokens.
We therefore study how this feedback changes the teacher's next-token distribution at positions where the judge names and defines evaluation criteria.

\paragraph{Per-position entropy shift.}
To measure how the language feedback $\pinfo$ changes next-token uncertainty at position $t$, we define
\begin{equation}
  \Delta H(s_t, \pinfo)
  \;=\;
  H\!\left(\pith(\cdot \mid s_t)\right)
  \;-\;
  H\!\left(\pith(\cdot \mid s_t, \pinfo)\right),
  \label{eq:ig-def}
\end{equation}
the entropy of the model's next-token distribution without $\pinfo$ minus the entropy with $\pinfo$, both evaluated at position $t$.
When the language feedback is fixed for an example, we abbreviate this as $\igt$.
\emph{Positive} $\igt$ means the teacher conditioned on $\pinfo$ has lower entropy than the student; $\pinfo$ has sharpened the teacher's distribution.
\emph{Negative} $\igt$ means the teacher has higher entropy than the student; $\pinfo$ has spread the teacher's distribution.
\emph{Near-zero} $\igt$ means little change in entropy, though not necessarily little change in the next-token distribution.

\paragraph{Both $\igt$ tails carry substantial distillation loss.}
We analyze 104,046 response positions from 102 \helpsteer{}~\citep{wang2025helpsteer3} validation rollouts generated by Qwen3-30B-A3B-Instruct-2507~\citep{yang2025qwen3}, with 34 examples each from code, general, and STEM.
Empirically, per-position reverse KL exhibits a U-shaped relationship with $\igt$: positions at either tail of the per-sequence $\igt$ distribution carry substantially more per-position KL than positions in the near-zero middle (Figure~\ref{fig:revkl-by-ig-tier}).
Partitioning each rollout into the bottom 30\%, middle 40\%, and top 30\% by $\igt$, both tails carry approximately 20--28$\times$ the per-position KL of the middle.
Thus, substantial distillation loss occurs both where the language feedback sharpens the teacher's distribution and where it spreads it.

\begin{figure}[!t]
  \centering
  \includegraphics[width=\linewidth]{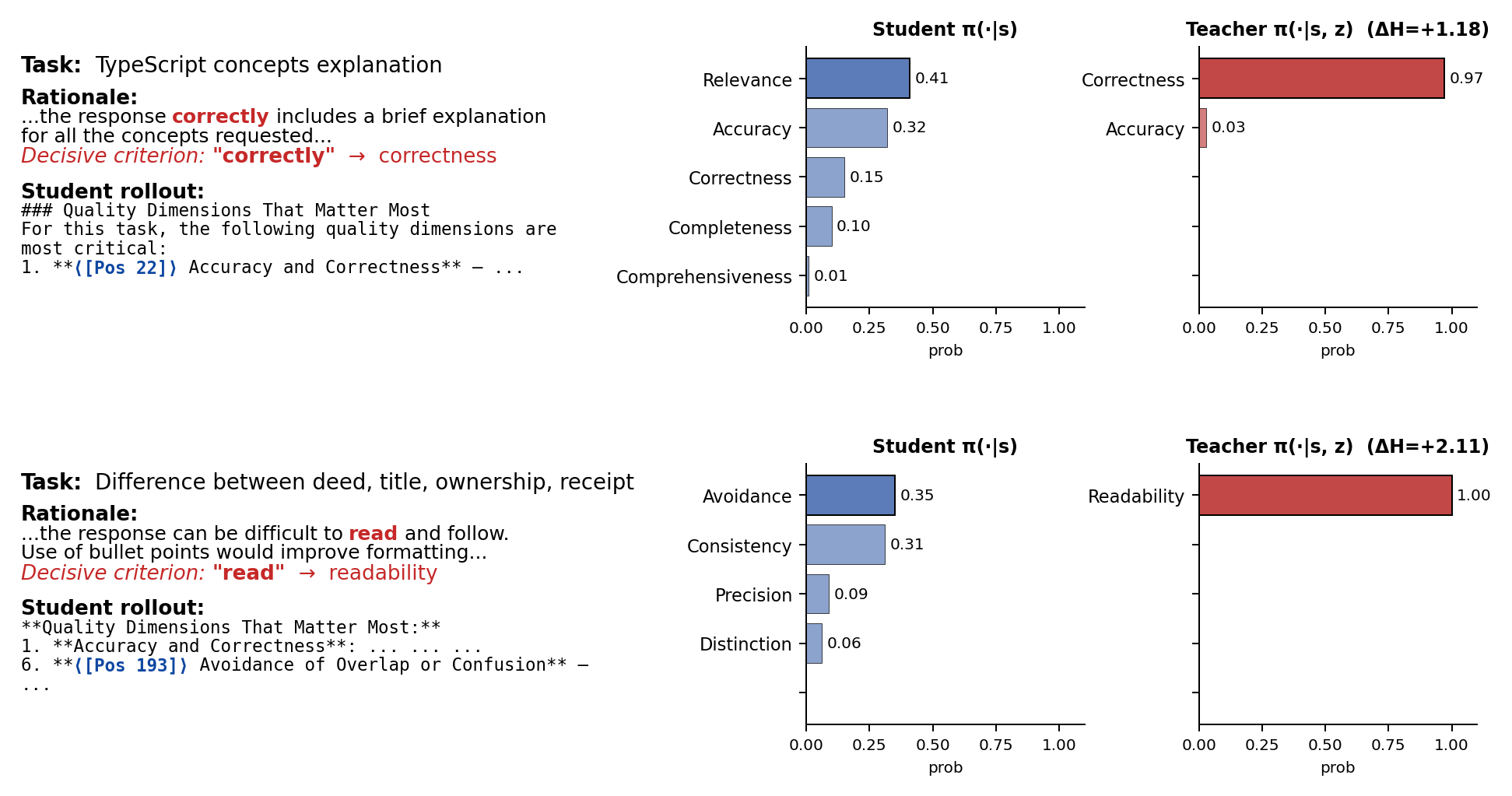}
  \caption{Context sharpening on two \helpsteer{} rollouts (Qwen3-30B-A3B-Instruct-2507). Each panel shows the user task, the rationale (anchor word \textcolor[HTML]{C62828}{\textbf{red}}), the student rollout excerpt with the annotated position $\langle$Pos~$N$$\rangle$ marked in \textcolor[HTML]{0D47A1}{\textbf{blue}}, and student vs.\ teacher top-5 next-token distributions at that position. All positions sit at criterion-naming slots in the rollout's opening evaluation-criteria list. The teacher concentrates probability on a token semantically equivalent to the rationale's anchor while the student is uncertain across multiple plausible criteria. Two further rollouts are shown in Figure~\ref{fig:sharpening-appx}.}
  \label{tab:sharpening-examples}
\end{figure}

\paragraph{Criterion-choice tokens account for disproportionate distillation loss.}
Criterion choice determines which evaluation criteria the judge invokes and how it weighs them, and is particularly important for subjective tasks.
Positions where the judge names and defines these criteria make this choice explicit in the intermediate trace.
Since the preference rationale identifies the decisive criteria, these positions provide a natural location to examine how language feedback shapes criterion choice.

We analyze the same 102 rollouts using GPT-5.5~\citep{openai2026gpt55}, with the generation and rationale annotated separately.
From the generation alone, the annotator marks \emph{criterion-selection spans}, comprising tokens that name and define evaluation criteria, and \emph{criterion-name positions}, marking each criterion's head noun.
These annotations are mapped to model-token positions. From the rationale alone, the annotator extracts the decisive criteria, which serve as the reference for the semantic analysis below.

Criterion-selection spans comprise only 16.7\% of response tokens but account for 44.4\% of total reverse-KL mass, corresponding to approximately 4.0$\times$ the per-token KL of other positions.
Criterion-name positions alone comprise 0.4\% of tokens but account for 11.1\% of total reverse-KL mass.
They also exhibit 5.7$\times$ larger mean $|\igt|$ than other response positions, and 94.8\% fall within the two extreme 30\% tails, which together contain 60\% of response positions. Thus, criterion-choice tokens account for disproportionate distillation loss, and criterion-name positions concentrate in both entropy-shift tails.
This motivates examining how the two tails differ in the criterion choices favored by the teacher.

\paragraph{Two regimes of language-feedback influence on criterion choice.}
The sign of $\igt$ distinguishes sharpening from spreading, but entropy alone does not reveal which criteria the teacher favors.
We therefore examine whether its candidate criterion names align with the decisive criteria extracted from the preference rationale.

Among the annotated criterion-name positions, we select those with $|\igt| \geq 0.50$, yielding 87 sharpening positions from 56 rollouts and 71 spreading positions from 51 rollouts.
At each position, we take the teacher's top-10 next-token candidates as candidate criterion-name tokens.
We force each token and greedily complete it into a criterion name, truncating at the head noun.
Using Qwen3-Embedding-8B~\citep{zhang2025qwen3embedding}, we score each completed name by its maximum cosine similarity to the extracted decisive criteria.
For sharpening, we score the name obtained from the top-1 token, measuring alignment of the teacher's most probable criterion.
For spreading, we average across all ten candidates, measuring alignment across the broader set of criteria.

As a control, we replace the teacher's language feedback with another example's rationale and repeat the candidate selection and completion procedure.
The student prefix, evaluation position, and regime assignment remain fixed as determined under the matched condition.
Both conditions are scored against the same decisive criteria from the original rationale.
We report paired mean differences, with 95\% confidence intervals obtained from 2,000 bootstrap resamples at the rollout level.

\begin{figure}[!t]
  \centering
  \includegraphics[width=\linewidth]{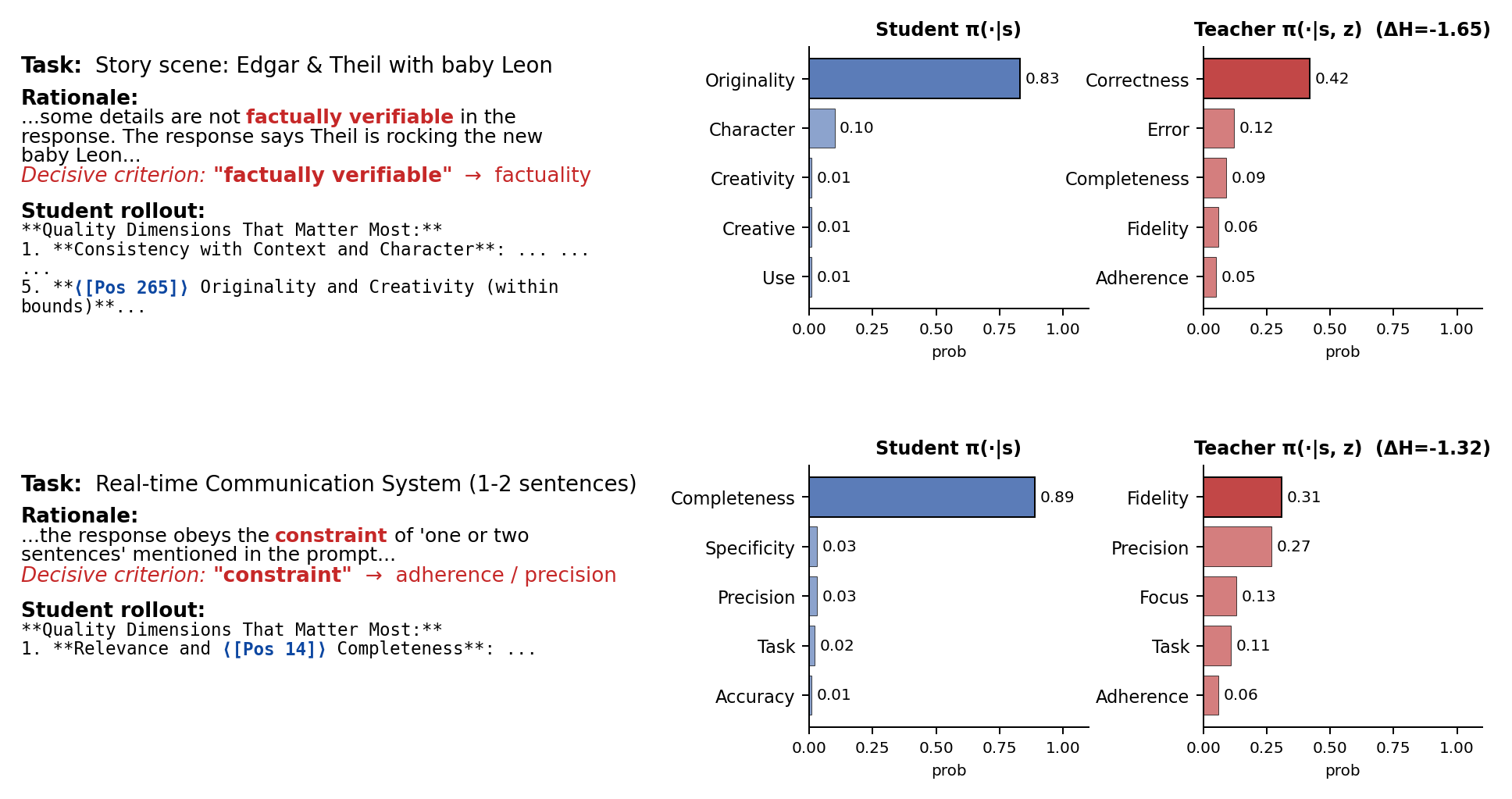}
  \caption{Context spreading on two \helpsteer{} rollouts (Qwen3-30B-A3B-Instruct-2507).
  Same layout as Figure~\ref{tab:sharpening-examples}.
  Each panel features a position where the student commits to a criterion that is non-decisive in the rationale (top-1 probability $\geq 0.79$), while the teacher reopens with a spread of alternatives that include semantic equivalents of the rationale's decisive criterion.
  All positions sit at criterion-naming slots in the rollout's opening evaluation-criteria list.
  Two further rollouts are shown in Figure~\ref{fig:spreading-appx}.}
  \label{tab:spreading-examples}
\end{figure}

\emph{Context sharpening.}
At positions with large positive $\igt$, the language feedback makes the teacher more confident than the student.
Figure~\ref{tab:sharpening-examples} illustrates how this concentrates probability on a specific decisive criterion.
In the TypeScript-explanation rollout, the rationale phrase \emph{``correctly includes''} causes the teacher to concentrate on \texttt{Correctness} (top-1 $0.97$), while the student is uncertain across several plausible criteria.
In the legal-term explanation rollout, \emph{``difficult to read''} causes the teacher to concentrate on \texttt{Readability} (top-1 $1.00$).

Across the annotated sharpening positions, the teacher's top-1 criterion name has mean similarity 0.713 to the decisive criteria under the matched rationale, compared with 0.647 under the control.
The paired difference is $+0.066$, with a 95\% confidence interval of $[0.033, 0.100]$.
Together with the lower teacher entropy, this supports the interpretation that context sharpening concentrates probability on a particular feedback-aligned criterion expression.
We interpret this concentrated supervision as encouraging memorization of a particular criterion expression rather than understanding of the underlying criterion.

\emph{Context spreading.}
At positions with large negative $\igt$, the language feedback makes the teacher less certain than the student.
Figure~\ref{tab:spreading-examples} illustrates how this reopens a criterion choice by spreading probability across alternatives.
In the story-writing rollout, the student commits to \texttt{Originality} (top-1 $0.83$), while the rationale's emphasis on \emph{``factually verifiable''} details shifts the teacher toward alternatives including \texttt{Correctness}, \texttt{Error}, and \texttt{Fidelity}.
In the project-description rollout, the student commits to \texttt{Completeness}, while the \emph{``one or two sentences''} constraint shifts the teacher toward alternatives including \texttt{Precision}, \texttt{Focus}, and \texttt{Adherence}.

Across the annotated spreading positions, mean similarity over the teacher's top-10 criterion names is 0.646 under the matched rationale, compared with 0.610 under the control.
The paired difference is $+0.037$, with a 95\% confidence interval of $[0.022, 0.051]$.
The result is consistent when using the top-3 or top-5 candidates.
Together with the higher teacher entropy, this supports the interpretation that context spreading distributes probability across multiple feedback-aligned alternatives.
We interpret this supervision as promoting semantic understanding of the underlying criterion by preserving multiple feedback-aligned alternatives.

\subsection{Position Masking Based on Entropy Shift}
\label{sec:ig}

Section~\ref{sec:theory} shows that context sharpening concentrates probability on a particular feedback-aligned criterion expression, while context spreading distributes probability across multiple feedback-aligned alternatives. Motivated by the possibility that preserving these alternatives promotes semantic understanding, we propose a per-generation mask that drops the upper tail and retains the lower tail of the entropy-shift distribution.
Concretely, we rank positions within each generation by $\igt$, mask the top $\rho$ fraction with the largest values, and retain the remaining bottom $1-\rho$ fraction:
\begin{equation}
\begin{aligned}
  \mathcal{L}_{\text{SD}}^{(\rho)}(\piS) \;&=\; \mathbb{E}_{(x, y_A, y_B, \pinfo) \sim \mathcal{D},\; \tau \sim \piS}
  \left[
    \frac{1}{\sum_t m_t}\sum_{t=1}^{T}
    m_t \cdot\mathrm{KL}\!\left(\piS(\cdot \mid s_t)\;\Big\|\;\mathrm{sg}\bigl[\piT(\cdot \mid s_t, \pinfo)\bigr]\right)
  \right], \\
  m_t \;&=\; \mathbf{1}\!\left[\igt \leq Q_{1-\rho}\right],
\end{aligned}
\label{eq:masked-sdpo-loss}
\end{equation}
where $Q_{1-\rho}$ is the $(1-\rho)$-quantile of $\{\ig_1, \dots, \ig_T\}$ for that generation.
The mask is detached from the computation graph.
Setting $\rho = 0$ recovers naive \sdpo{}.

\paragraph{Entropy-shift masking is associated with broader criterion diversity.}
\begin{wrapfigure}{r}{0.42\linewidth}
  \vspace{-1.0\baselineskip}
  \centering
  \includegraphics[width=\linewidth]{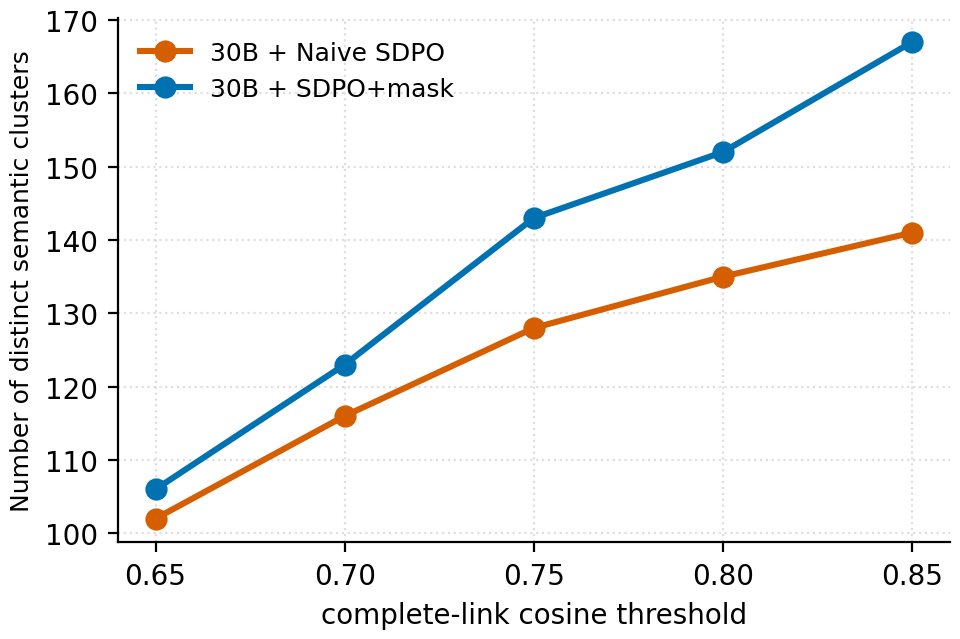}
  \caption{Distinct semantic clusters of evaluation criteria invoked by trained 30B judges across a sweep of complete-link cosine clustering thresholds. Naive \sdpo{} invokes a markedly smaller pool than \sdpo{}+mask at every threshold; the gap widens with stricter clustering.}
  \label{fig:criteria-pool-sweep}
  \vspace{-1.0\baselineskip}
\end{wrapfigure}
Section~\ref{sec:theory} shows that, at the analyzed criterion-name positions, context spreading distributes probability across multiple feedback-aligned alternatives, whereas context sharpening concentrates probability on a particular criterion expression. This suggests that prioritizing lower-entropy-shift positions during distillation helps preserve a broader repertoire of evaluation criteria. We examine this possibility by measuring the diversity of criteria invoked by the trained judges at inference time. For each judgment a trained judge produces (across 300 samples each from \rmbench{}~\citep{liu2025rmbench}, \rewardbench{}~\citep{malik2025rewardbenchv2}, and \helpsteer{} validation~\citep{wang2025helpsteer3}), we extract the evaluation criteria the model proposes and uses (e.g., \textit{``adherence to user intent''}, \textit{``factual accuracy''}, \textit{``relevance to the request''}), apply text normalization (lowercase, strip punctuation, sort tokens to merge order-variants), and embed each criterion string with Qwen3-Embedding-8B~\citep{zhang2025qwen3embedding}. We then cluster the resulting 4096-dimensional vectors with \emph{complete-link agglomerative clustering} at cosine threshold $t$: two criterion strings belong to the same cluster only if \emph{every pair} within the cluster has cosine similarity $\geq t$. This procedure groups semantically similar criterion expressions (e.g., \textit{``factual accuracy,''} \textit{``factual correctness,''} and \textit{``accuracy of facts''}), reducing sensitivity to differences in wording. Larger $t$ enforces tighter clusters; lower $t$ allows looser merging.

Figure~\ref{fig:criteria-pool-sweep} shows that \sdpo{}+mask invokes more distinct criterion clusters than naive \sdpo{} at every evaluated clustering threshold, with the difference increasing from 4 clusters at $t=0.65$ to 26 at $t=0.85$. This pattern is consistent with the interpretation suggested by \S\ref{sec:theory}: supervision that preserves multiple feedback-aligned alternatives helps maintain a broader criterion repertoire after training. Different prompts call for different evaluation criteria, making criterion diversity a relevant property of a general-purpose judge.

%% file: sections/experiments.tex
\section{Experiments}
\label{sec:experiments}

\subsection{Setup}
\label{sec:experiments-setup}

\paragraph{Base models.}
We train two instruct models as judges: Qwen3-4B-Instruct-2507\footnote{\url{https://huggingface.co/Qwen/Qwen3-4B-Instruct-2507}} and Qwen3-30B-A3B-Instruct-2507\footnote{\url{https://huggingface.co/Qwen/Qwen3-30B-A3B-Instruct-2507}}.
The judge prompt template is given in Appendix~\ref{app:prompt}; we tuned a small set of candidate templates on the base model and chose the one that yielded the best validation accuracy.

\paragraph{Training data.}
All methods are trained on a cleaned version of \helpsteer{}~\citep{wang2025helpsteer3} with three modifications applied in order:
(1)~filter out rows with \texttt{domain == "multilingual"} or \texttt{overall\_preference == 0} (tie);
(2)~within-split deduplication by content hash (\texttt{sha1(context, response1, response2)}), keeping the first occurrence; upstream \helpsteer{} contains $\sim$35\% byte-identical duplicate rows after step~(1);
(3)~cross-split deduplication: drop validation rows whose content hash also appears in train-split; upstream \helpsteer{} has $\sim$918 of $\sim$1{,}553 filtered-deduped validation rows that are byte-identical to a train-split row and would otherwise cause train/val leakage.
Each retained example contains a prompt, two candidate responses, a gold preference label, and a one- or two-sentence human-written rationale explaining the label.

\paragraph{Language feedback types.}

We mainly use the preference rationale as the source of language feedback $\pinfo$. Preference rationales are the most naturally available and easiest-to-collect form of textual feedback during preference-data construction. To assign a binary preference label, an annotator must already compare the two responses and identify why one is better. To test sensitivity to a substantially different language-feedback format, we also generated a per-example rubric for every \helpsteer{} sample using Claude Opus 4.8~\citep{anthropic2026opus48}, following the iterative rubric-generation procedures of \citep{shen2026rethinkingrubric}. Generating rubrics requires a separate generation or annotation process and careful design to ensure that the criteria are discriminative, non-redundant, and aligned with the response pair and preference direction. The generated rubrics average 2,062 characters, making them approximately 7.7 times longer than the original preference rationales. Results using these rubrics as language feedback are reported in Appendix~\ref{app:rubric-feedback}.

\paragraph{Evaluation benchmarks.}
We evaluate out-of-distribution (OOD) judge accuracy on \rmbench{}~\citep{liu2025rmbench} and \rewardbench{}~\citep{malik2025rewardbenchv2}.

%

\paragraph{Baselines.}
We compare three on-policy training methods. We use \drgrpo{} as the outcome-supervised RL baseline. Naive \sdpo{} ($\rho = 0$) is the objective in Eq.~\ref{eq:sdpo-loss} with all positions included. \sdpoig{} uses $\rho = 0.7$, masking the top 70\% of positions by $\igt$ and keeping the 30\% of positions with the lowest entropy shifts. Implementation details for all three are in Appendix~\ref{app:implementation}. We additionally compare against strong prompting baselines (DeepSeek-R1~\citep{deepseekai2025r1}, Claude-Sonnet-4~\citep{anthropic2025claude4}) and trained reward-model baselines (Think-RM-8B~\citep{hong2025thinkrm}, RM-R1-DeepSeek-Distilled-Qwen-7B and RM-R1-DeepSeek-Distilled-Qwen-32B~\citep{chen2026rmr1}, Llama-3.3-Nemotron-Super-49B-GenRM~\citep{nvidia2025nemotrongenrm}) evaluated from their authors' released checkpoints, and RationaleRM-30B~\citep{wang2026outcome} for which we cite the authors' reported numbers.


\begin{table}[!t]
  \caption{Judge accuracy (\%) on \rmbench{} and \rewardbench{}. Dark gray (in bold) and light gray highlight the best and second-best performance per column, respectively. RationaleRM-30B numbers are taken from \citet{wang2026outcome} as the model has not been released; \rewardbench{} cells are left empty.}
  \label{tab:main-results}
  \centering
  \scriptsize
  \setlength{\tabcolsep}{3pt}
  \resizebox{\linewidth}{!}{%
  \begin{tabular}{lcccccccccccc}
    \toprule
    \multirow{2}{*}{Models} & \multicolumn{5}{c}{\rmbench{}} & \multicolumn{6}{c}{\rewardbench{}} & \multirow{2}{*}{Total Avg.} \\
    \cmidrule(lr){2-6} \cmidrule(lr){7-12}
     & Chat & Code & Math & Safety & Overall & Factuality & Focus & Math & Precise IF & Safety & Overall & \\
    \midrule
    \multicolumn{13}{l}{\textit{Prompting}} \\
    Claude-Sonnet-4-20250514 & 79.93 & 81.77 & 93.05 & 95.34 & 87.52 & 84.00 & 87.27 & 85.25 & 56.88 & 94.00 & 81.48 & 84.50 \\
    DeepSeek-R1   & 78.38 & 79.29 & 94.01 & 92.27 & 85.99 & 74.53 & 80.40 & 87.43 & 43.12 & 86.00 & 74.30 & 80.15 \\
    \midrule
    \multicolumn{13}{l}{\textit{Training (Small Model $<$10B)}} \\
    Think-RM-8B & 65.20 & 54.63 & 71.92 & \snd{91.71} & 70.87 & 49.26 & 62.42 & 58.47 & 26.88 & 75.56 & 54.52 & 62.70 \\
    RM-R1-DeepSeek-Distilled-Qwen-7B & 64.51 & 60.87 & 88.55 & 85.99 & 74.98 & 28.21 & 58.59 & 74.32 & 19.38 & 51.56 & 46.41 & 60.70 \\
    Qwen3-4B-Instruct + \drgrpo{}           & 68.91 & \snd{73.83} & \snd{94.01} & 90.93 & 81.92 & 62.74 & 75.76 & \best{86.34} & \snd{43.12} & 83.78 & 70.35 & 76.14 \\
    Qwen3-4B-Instruct + Naive \sdpo{}       & \snd{75.71} & 73.34 & 93.51 & \best{92.06} & \snd{83.66} & \snd{68.42} & \best{80.20} & \snd{85.79} & 38.75 & \snd{85.56} & \snd{71.74} & \snd{77.70} \\
    Qwen3-4B-Instruct + \sdpo{}+mask & \best{77.95} & \best{75.88} & \best{94.10} & 90.95 & \best{84.72} & \best{70.74} & \snd{77.78} & 84.15 & \best{45.00} & \best{89.11} & \best{73.36} & \best{79.04} \\
    \midrule
    \multicolumn{13}{l}{\textit{Training (Medium and Large Model $\geq$10B)}} \\
    RM-R1-DeepSeek-Distilled-Qwen-32B & 72.78 & 76.12 & 92.46 & \best{94.89} & 84.06 & 60.84 & 80.40 & \snd{85.25} & 38.75 & 82.89 & 69.63 & 76.85 \\
    RationaleRM-30B & 74.90 & \best{84.40} & 95.50 & 93.60 & \snd{87.10} & -- & -- & -- & -- & -- & -- & -- \\
    Llama-3.3-Nemotron-Super-49B-GenRM & 73.64 & 73.34 & 90.55 & 90.15 & 81.92 & 70.95 & 81.41 & 83.61 & 44.38 & 81.56 & 72.38 & 77.15 \\
    Qwen3-30B-A3B-Instruct + \drgrpo{}           & 74.07 & 79.68 & \snd{95.59} & 93.78 & 85.78 & 71.79 & 80.00 & \best{87.43} & \snd{42.50} & \best{92.44} & \snd{74.83} & \snd{80.31} \\
    Qwen3-30B-A3B-Instruct + Naive \sdpo{}       & \snd{76.74} & 76.56 & 94.45 & 94.26 & 85.50 & \snd{78.53} & \best{86.26} & 81.42 & 40.62 & 87.33 & \snd{74.83} & 80.17 \\
    Qwen3-30B-A3B-Instruct + \sdpo{}+mask & \best{80.53} & \snd{80.07} & \best{95.63} & \snd{94.81} & \best{87.76} & \best{79.16} & \snd{85.66} & 84.15 & \best{48.75} & \snd{88.44} & \best{77.23} & \best{82.50} \\
    \bottomrule
  \end{tabular}%
  }
\end{table}

\subsection{Main Results}
\label{sec:main-results}

Table~\ref{tab:main-results} reports judge accuracy on \rmbench{} and \rewardbench{} for the Qwen3-4B-Instruct and Qwen3-30B-A3B-Instruct base models, using preference rationales as language feedback, alongside prompting-based and trained reward-model baselines.

\paragraph{Main findings.}
Per-category results in Table~\ref{tab:main-results} can be split along the objective/subjective axis.

\emph{Outcome-supervised RL vs.\ self-distillation.}
On the objective subcategories (e.g., math, coding), where the decisive criterion is clear-cut, \drgrpo{} sharpens application rigor against that criterion and is competitive with both naive \sdpo{} and \sdpo{}+mask: at 30B it matches them on RM-Bench Math (95.59 (\drgrpo{}) vs 94.45 (\sdpo{}) / 95.63 (\sdpo{}+mask)) and beats them on \rewardbench{} Math (87.43 vs 81.42 / 84.15).
On the subjective subcategories (e.g., chat helpfulness; factuality, where the judge must prioritize factual accuracy over otherwise persuasive presentation; focus, which tests detection of high-quality, on-topic answers to general user queries), the gap flips: both \sdpo{} variants gain over \drgrpo{} by 2--9 percentage points (30B Chat 74.07 vs 76.74 / 80.53; 30B Factuality 71.79 vs 78.53 / 79.16; 30B Focus 80.00 vs 86.26 / 85.66; 4B Chat 68.91 vs 75.71 / 77.95; 4B Factuality 62.74 vs 68.42 / 70.74; 4B Focus 75.76 vs 80.20 / 77.78).

\emph{Naive \sdpo{} vs.\ \sdpo{}+mask.}
\sdpo{}+mask consistently improves over naive \sdpo{} on overall accuracy at both scales (Total Avg: $77.70 \to 79.04$ at 4B; $80.17 \to 82.50$ at 30B), with the largest per-subcategory gains spread across different subcategories (30B RM-Bench Code $76.56 \to 80.07$; 30B Chat $76.74 \to 80.53$; 30B \rewardbench{} Precise IF $40.62 \to 48.75$). Because both benchmarks are OOD with respect to the training data, these gains are consistent with the proposed mechanism: masking higher-entropy-shift positions reduces overreliance on particular criterion expressions, while the broader inference-time criterion vocabulary in Figure~\ref{fig:criteria-pool-sweep} provides additional support for this interpretation.

Our 30B \sdpo{}+mask judge outperforms leading judge baselines in the literature, including RM-R1-DeepSeek-Distilled-Qwen-32B, Llama-3.3-Nemotron-Super-49B-GenRM, and RationaleRM-30B, on overall benchmark accuracy, and remains comparable to Claude-Sonnet-4.

\subsection{Downstream Utility for Policy Optimization}
\label{sec:downstream-selection}

The benchmark results above evaluate judges directly, but a reward model is ultimately used to determine which on-policy outputs receive higher reward and advantage during policy optimization.
The strongest downstream validation would train separate policies with GRPO using each learned judge as the reward model.
Because this requires multiple full RLHF runs, we instead evaluate the core pairwise-selection operation used by judge-guided GRPO through a controlled multi-round best-of-eight tournament.

\paragraph{Setup.}
For each subjective creative-writing prompt from Arena-Hard-v2~\citep{li2024arenahard}, GLM-4.5-Air~\citep{zeng2025glm45} generates eight responses with temperature $1.0$, top-$p$ $1.0$, and a maximum of 8K new tokens.
We randomly pair the eight responses into four matchups and use a judge to select the winner of each pair.
We repeat this procedure with the four winners, and then with the two remaining responses, until one final response is selected.
We run the tournament separately using Qwen3-30B-A3B-Instruct judges trained with \drgrpo{}, naive \sdpo{}, and \sdpo{}+mask ($\rho=0.7$).
Claude Sonnet 5~\citep{anthropic2026sonnet5} evaluates the selected responses against the Arena-Hard-v2 reference responses to compute the creative-writing win rate.

\begin{table}[t]
  \captionsetup{skip=5pt}
  \caption{Creative-writing win rate of responses selected through multi-round best-of-eight tournaments. Each tournament uses a different Qwen3-30B-A3B-Instruct judge as its pairwise selector; selected responses are evaluated against the Arena-Hard-v2 reference responses by Claude Sonnet 5.}
  \label{tab:downstream-selection}
  \centering
  \small
  \begin{tabular}{lc}
    \toprule
    Selector & Creative-writing win rate \\
    \midrule
    \drgrpo{} & 0.556 \\
    Naive \sdpo{} & \snd{0.612} \\
    \sdpo{}+mask ($\rho=0.7$) & \best{0.632} \\
    \bottomrule
  \end{tabular}
\end{table}

\sdpo{}+mask selects the strongest downstream responses, improving the creative-writing win rate over the \drgrpo{}-trained judge by $7.6$ percentage points and over naive \sdpo{} by $2.0$ percentage points.
Although this tournament does not include the subsequent gradient-based policy updates of a full GRPO run, it directly tests the repeated pairwise reward comparisons that determine which on-policy outputs would be preferentially reinforced.
The result therefore provides evidence that the gains from \sdpo{}+mask are not confined to standalone judge benchmarks: when used as a reward selector, it more reliably favors responses preferred under the downstream evaluation.

\subsection{Ablations}
\label{sec:ablations}

%
%

\subsubsection{Mask-fraction ($\rho$) sweep}
\label{sec:rho-sweep}

\begin{table}[!h]
  \caption{$\rho$-sweep on Qwen3-30B-A3B-Instruct, all masking the top-$\rho$ fraction by $\igt$ and training on the remaining bottom $(1-\rho)$.}
  \label{tab:rho-sweep}
  \centering
  \scriptsize
  \setlength{\tabcolsep}{3pt}
  \resizebox{\linewidth}{!}{%
  \begin{tabular}{lcccccccccccc}
    \toprule
    \multirow{2}{*}{$\rho$ (fraction masked)} & \multicolumn{5}{c}{\rmbench{}} & \multicolumn{6}{c}{\rewardbench{}} & \multirow{2}{*}{Total Avg.} \\
    \cmidrule(lr){2-6} \cmidrule(lr){7-12}
     & Chat & Code & Math & Safety & Overall & Factuality & Focus & Math & Precise IF & Safety & Overall & \\
    \midrule
    $0.00$ (naive \sdpo{})  & 76.74 & 76.56 & 94.45 & \snd{94.26} & 85.50 & 78.53 & \best{86.26} & 81.42 & 40.62 & \snd{87.33} & 74.83 & 80.17 \\
    $0.30$                  & \snd{77.52} & 75.78 & \snd{95.02} & 94.21 & 85.63 & \best{80.00} & 83.03 & \best{85.79} & 43.75 & 85.56 & \snd{75.63} & 80.63 \\
    $0.50$                  & 77.17 & \snd{77.49} & 94.85 & 94.15 & \snd{85.92} & 78.95 & 83.43 & \snd{84.15} & \snd{46.88} & 83.78 & 75.44 & \snd{80.68} \\
    $0.70$ (canonical)      & \best{80.53} & \best{80.07} & \best{95.63} & \best{94.81} & \best{87.76} & \snd{79.16} & \snd{85.66} & \snd{84.15} & \best{48.75} & \best{88.44} & \best{77.23} & \best{82.50} \\
    \bottomrule
  \end{tabular}%
  }
\end{table}

We sweep the mask fraction $\rho \in \{0, 0.3, 0.5, 0.7\}$ on Qwen3-30B-A3B-Instruct (Table~\ref{tab:rho-sweep}) to characterize how aggressive the mask should be; the corresponding Qwen3-4B-Instruct results are reported in Appendix~\ref{app:additional-ablations}.
$\rho=0$ recovers naive \sdpo{} (no positions masked).
Overall accuracy rises as we mask more of the upper $\igt$ tail (Total Avg: $80.17 \to 80.63 \to 80.68 \to 82.50$ across $\rho \in \{0, 0.3, 0.5, 0.7\}$), with $\rho = 0.7$ achieving the highest overall accuracy among the tested mask fractions.
The same pattern holds at 4B, where $\rho=0.7$ also obtains the highest total average (Table~\ref{tab:rho-sweep-4b}).

\subsubsection{Selector ablation}
\label{sec:selector-ablation}

\begin{table}[!h]
  \caption{Selector ablation on Qwen3-30B-A3B-Instruct, all at matched mask fraction $\rho = 0.7$.}
  \label{tab:selector-ablation}
  \centering
  \scriptsize
  \setlength{\tabcolsep}{3pt}
  \resizebox{\linewidth}{!}{%
  \begin{tabular}{lcccccccccccc}
    \toprule
    \multirow{2}{*}{Selector} & \multicolumn{5}{c}{\rmbench{}} & \multicolumn{6}{c}{\rewardbench{}} & \multirow{2}{*}{Total Avg.} \\
    \cmidrule(lr){2-6} \cmidrule(lr){7-12}
     & Chat & Code & Math & Safety & Overall & Factuality & Focus & Math & Precise IF & Safety & Overall & \\
    \midrule
    Mask top $\rho$ fraction by $\igt$ (ours; drops highest-shift positions)         & \best{80.53} & \best{80.07} & \best{95.63} & \best{94.81} & \best{87.76} & \best{79.16} & 85.66 & 84.15 & \best{48.75} & \best{88.44} & \best{77.23} & \best{82.50} \\
    Mask bottom $\rho$ fraction by $\igt$ (direction flip; drops lowest-shift positions) & 75.37 & 74.22 & 93.40 & 94.13 & 84.28 & 72.63 & 84.95 & \best{85.55} & 40.67 & \snd{87.05} & 74.17 & 79.23 \\
    Mask bottom $\rho$ fraction by $|\igt|$ (both tails; drops smallest absolute shifts) & \snd{76.14} & \snd{76.17} & 93.47 & 93.78 & \snd{84.89} & 73.05 & \best{86.60} & 84.39 & \snd{44.00} & 84.77 & 74.56 & \snd{79.73} \\
    Random masking & 75.11 & 75.88 & 93.72 & \snd{94.41} & 84.78 & 77.47 & 85.45 & 81.97 & 40.00 & 86.67 & 74.31 & 79.55 \\
    Mask bottom $\rho$ fraction by student entropy & 74.59 & 75.44 & 93.59 & 94.15 & 84.44 & 76.63 & \snd{86.46} & \snd{84.70} & 40.00 & 85.78 & \snd{74.71} & 79.57 \\
    Mask bottom $\rho$ fraction by teacher entropy & 75.45 & 75.44 & \snd{94.01} & 93.35 & 84.56 & \snd{78.11} & 83.84 & 83.06 & 35.62 & 82.67 & 72.66 & 78.61 \\
    \bottomrule
  \end{tabular}%
  }
\end{table}

Is the OOD gain specific to the entropy-shift criterion, or does any sensible per-position selector at matched $\rho$ give the same benefit?
We compare our top-$\igt$-masked selector against five alternatives at matched $\rho = 0.7$ on Qwen3-30B-A3B-Instruct (Table~\ref{tab:selector-ablation}):
\begin{itemize}
  \item $\igt$, mask bottom $\rho$ fraction (\emph{direction flip}): drop the positions with the lowest entropy shifts and retain the highest-shift positions instead. Tests whether the gain depends on the masking direction.
  \item $|\igt|$, mask bottom $\rho$ fraction (\emph{both tails}): drop positions with near-zero entropy shifts and retain positions with large absolute entropy shifts, where the teacher and student entropies differ substantially. Tests whether retaining positions with large absolute entropy shifts is sufficient, regardless of whether the shift is positive or negative.
  \item Random masking: mask a uniformly sampled $\rho$ fraction of response positions. Tests whether the gain follows simply from reducing the number of supervised positions.
  \item Student-entropy masking: mask the bottom $\rho$ fraction of positions by student next-token entropy. Tests whether retaining positions where the unconditioned student is uncertain is sufficient.
  \item Teacher-entropy masking: mask the bottom $\rho$ fraction of positions by teacher next-token entropy. Tests whether retaining positions where the feedback-conditioned teacher is uncertain is sufficient.
\end{itemize}
Our top-$\igt$-masked selector outperforms all five controls on overall \rmbench{} and \rewardbench{} accuracy. The gain is therefore not explained by the masking fraction alone, by either distribution's entropy in isolation, by reversing the masking direction, or by retaining high-$|\igt|$ positions of either sign. Together, these ablations support selecting positions by signed entropy shift, with lower-shift selection outperforming the tested alternatives.
The corresponding Qwen3-4B-Instruct results are reported in Appendix~\ref{app:additional-ablations}.

%% file: sections/conclusion.tex
\section{Conclusion}
\label{sec:conclusion}

We studied on-policy dense supervision for training LLM judges on subjective tasks, where the verdict hinges on which evaluation criteria the judge invokes and how it weighs them. Outcome-supervised RL credits every token by the final verdict and does not provide explicit guidance on criterion choice; \sdpo{} converts the per-example language feedback that names the decisive criterion into per-position supervision, but not every position carries equally useful signal. Using the per-position entropy shift between teacher and student, we identified two regimes, context sharpening and context spreading, and proposed a simple mask that retains the lower tail of the entropy-shift distribution. Across model sizes, \sdpo{} outperforms outcome-supervised RL on the evaluated subjective tasks, and our mask further improves generalization over naive \sdpo{}.

%% file: sections/limitations.tex
\section{Limitations and Scope}
\label{sec:limitations}

Our method has two main limitations. First, the method depends on feedback quality: entropy-shift masking does not guarantee useful supervision when the feedback fails to identify a decisive criterion or provides only generic guidance. Second, our method does not explicitly address known limitations of self-distillation, including hallucination and training instability in long-chain-of-thought reasoning models~\citep{kim2026selfdistillation}. We focus on instruction-tuned models to support efficient judge inference; extending the method to long-chain-of-thought reasoning models remains future work.

%% file: sections/appendix.tex
\section{Additional Context-Sharpening and Context-Spreading Examples}
\label{app:additional-examples}

\begin{figure}[h]
  \centering
  \includegraphics[width=\linewidth]{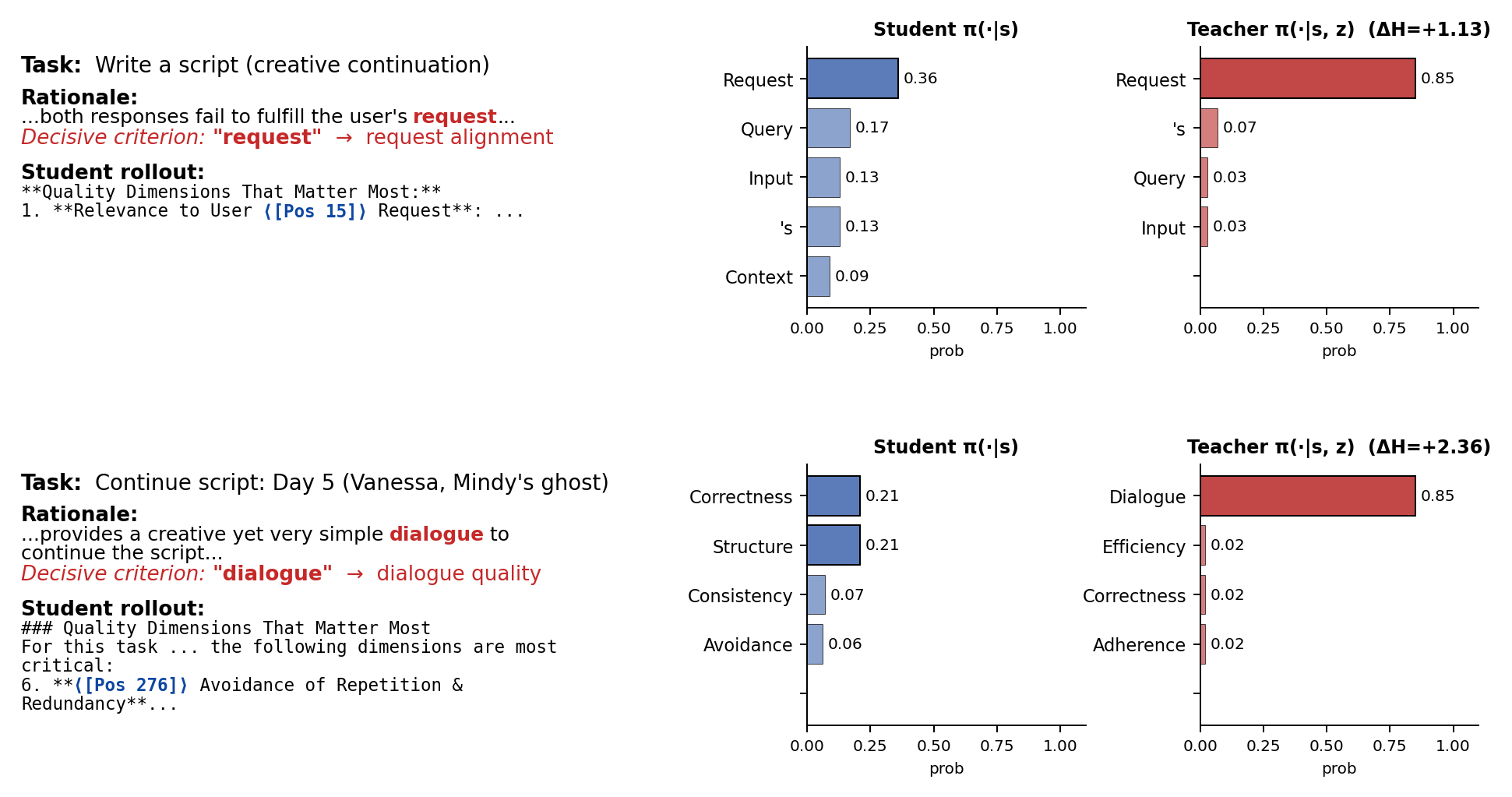}
  \caption{Additional context-sharpening examples on two \helpsteer{} rollouts (Qwen3-30B-A3B-Instruct-2507).}
  \label{fig:sharpening-appx}
\end{figure}

\begin{figure}[h]
  \centering
  \includegraphics[width=\linewidth]{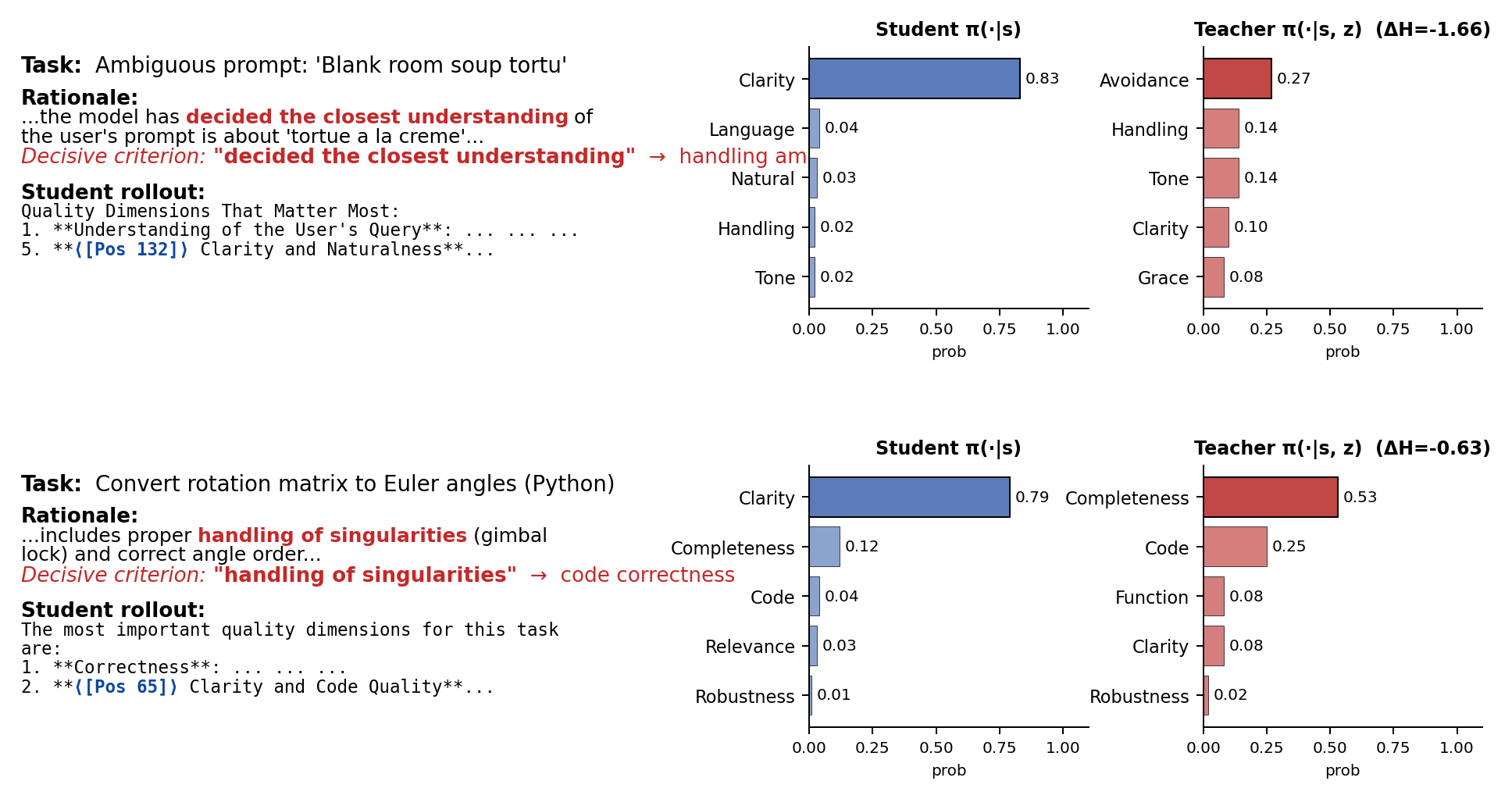}
  \caption{Additional context-spreading examples on two \helpsteer{} rollouts (Qwen3-30B-A3B-Instruct-2507).}
  \label{fig:spreading-appx}
\end{figure}

Figures~\ref{fig:sharpening-appx} and~\ref{fig:spreading-appx} present two further rollouts each for context sharpening and context spreading, supplementing the two featured examples in the main body (Figures~\ref{tab:sharpening-examples} and~\ref{tab:spreading-examples}). The plot layout and construction are identical to the main-body figures.

\section{Additional Experiments with Rubric Feedback}
\label{app:rubric-feedback}

We repeat the Qwen3-4B-Instruct self-distillation experiments using the generated per-example rubrics instead of the original preference rationales as language feedback $\pinfo$.
Table~\ref{tab:rubric-results} compares naive \sdpo{} and \sdpo{}+mask under both feedback formats, along with the same \drgrpo{} outcome-supervised baseline reported in the main results.

\begin{table}[h]
  \caption{Judge accuracy (\%) for Qwen3-4B-Instruct using preference rationales or generated per-example rubrics as language feedback. Dark gray (in bold) and light gray highlight the best and second-best performance per column, respectively.}
  \label{tab:rubric-results}
  \centering
  \scriptsize
  \setlength{\tabcolsep}{3pt}
  \resizebox{\linewidth}{!}{%
  \begin{tabular}{lcccccccccccc}
    \toprule
    \multirow{2}{*}{Models} & \multicolumn{5}{c}{\rmbench{}} & \multicolumn{6}{c}{\rewardbench{}} & \multirow{2}{*}{Total Avg.} \\
    \cmidrule(lr){2-6} \cmidrule(lr){7-12}
     & Chat & Code & Math & Safety & Overall & Factuality & Focus & Math & Precise IF & Safety & Overall & \\
    \midrule
    Qwen3-4B-Instruct + \drgrpo{}
      & 68.91 & 73.83 & \snd{94.01} & 90.93 & 81.92
      & 62.74 & 75.76 & \best{86.34} & 43.12 & 83.78 & 70.35 & 76.14 \\
    Qwen3-4B-Instruct + Naive \sdpo{} (rubric)
      & 75.45 & 75.19 & 93.57 & \best{92.34} & 84.14
      & 61.68 & 74.23 & 84.39 & 36.00 & 79.09 & 67.08 & 75.61 \\
    Qwen3-4B-Instruct + \sdpo{}+mask (rubric)
      & \snd{77.26} & \best{75.97} & 93.28 & 91.91 & \snd{84.61}
      & 65.89 & 75.05 & 84.97 & \best{48.00} & 79.55 & 70.69 & 77.65 \\
    Qwen3-4B-Instruct + Naive \sdpo{} (rationale)
      & 75.71 & 73.34 & 93.51 & \snd{92.06} & 83.66
      & \snd{68.42} & \best{80.20} & \snd{85.79} & 38.75 & \snd{85.56} & \snd{71.74} & \snd{77.70} \\
    Qwen3-4B-Instruct + \sdpo{}+mask (rationale)
      & \best{77.95} & \snd{75.88} & \best{94.10} & 90.95 & \best{84.72}
      & \best{70.74} & \snd{77.78} & 84.15 & \snd{45.00} & \best{89.11} & \best{73.36} & \best{79.04} \\
    \bottomrule
  \end{tabular}%
  }
\end{table}

Entropy-shift masking improves the total average under both feedback formats: from 75.61 to 77.65 with generated rubrics and from 77.70 to 79.04 with preference rationales. This suggests that the benefit of entropy-shift masking is not specific to one language-feedback format.

Second, the rationale-based configuration achieves stronger overall performance than the rubric-based configuration, despite using substantially shorter feedback. One possible explanation is that the original rationales focus directly on the decisive reasons for the observed preferences, whereas generated rubrics introduce additional criteria that are less relevant to those preferences.

These results support our practical choice of preference rationales: they are the most naturally available and easiest-to-collect textual feedback during preference-data construction, require little additional annotation beyond the preference judgment itself, and outperform the substantially longer, separately generated rubrics in our experiments.

\section{Additional Ablation Results}
\label{app:additional-ablations}

Table~\ref{tab:rho-sweep-4b} reports the full mask-fraction sweep for Qwen3-4B-Instruct, complementing the Qwen3-30B-A3B-Instruct results in Table~\ref{tab:rho-sweep}.

\begin{table}[h]
  \caption{$\rho$-sweep on Qwen3-4B-Instruct, all masking the top-$\rho$ fraction by $\igt$ and training on the remaining bottom $(1-\rho)$.}
  \label{tab:rho-sweep-4b}
  \centering
  \scriptsize
  \setlength{\tabcolsep}{3pt}
  \resizebox{\linewidth}{!}{%
  \begin{tabular}{lcccccccccccc}
    \toprule
    \multirow{2}{*}{$\rho$ (fraction masked)} & \multicolumn{5}{c}{\rmbench{}} & \multicolumn{6}{c}{\rewardbench{}} & \multirow{2}{*}{Total Avg.} \\
    \cmidrule(lr){2-6} \cmidrule(lr){7-12}
     & Chat & Code & Math & Safety & Overall & Factuality & Focus & Math & Precise IF & Safety & Overall & \\
    \midrule
    $0.00$ (naive \sdpo{})  & 75.71 & 73.34 & 93.51 & \best{92.06} & 83.66 & \snd{68.42} & \best{80.20} & \best{85.79} & 38.75 & 85.56 & 71.74 & 77.70 \\
    $0.30$                  & 75.54 & 74.51 & \snd{93.97} & 91.38 & 83.85 & 67.79 & 75.96 & 83.06 & 41.88 & 87.33 & 71.20 & 77.53 \\
    $0.50$                  & \snd{76.92} & \snd{75.54} & 93.82 & \snd{91.74} & \snd{84.50} & 67.58 & 76.36 & 83.61 & \best{46.88} & \snd{88.22} & \snd{72.53} & \snd{78.52} \\
    $0.70$ (canonical)      & \best{77.95} & \best{75.88} & \best{94.10} & 90.95 & \best{84.72} & \best{70.74} & 77.78 & 84.15 & \snd{45.00} & \best{89.11} & \best{73.36} & \best{79.04} \\
    \bottomrule
  \end{tabular}%
  }
\end{table}

At 4B, the total average changes from 77.70 with naive \sdpo{} to 77.53, 78.52, and 79.04 as $\rho$ increases to $0.3$, $0.5$, and $0.7$, respectively. Thus, the canonical $\rho=0.7$ setting is also the strongest configuration at the smaller model scale.

Table~\ref{tab:selector-ablation-4b} reports the corresponding selector ablation at 4B. As in the 30B results, every selector uses the same mask fraction, $\rho=0.7$.

\begin{table}[h]
  \caption{Selector ablation on Qwen3-4B-Instruct, all at matched mask fraction $\rho=0.7$.}
  \label{tab:selector-ablation-4b}
  \centering
  \scriptsize
  \setlength{\tabcolsep}{3pt}
  \resizebox{\linewidth}{!}{%
  \begin{tabular}{lcccccccccccc}
    \toprule
    \multirow{2}{*}{Selector} & \multicolumn{5}{c}{\rmbench{}} & \multicolumn{6}{c}{\rewardbench{}} & \multirow{2}{*}{Total Avg.} \\
    \cmidrule(lr){2-6} \cmidrule(lr){7-12}
     & Chat & Code & Math & Safety & Overall & Factuality & Focus & Math & Precise IF & Safety & Overall & \\
    \midrule
    Mask top $\rho$ fraction by $\igt$ (ours; drops highest-shift positions) & \best{77.95} & \best{75.88} & \best{94.10} & 90.95 & \best{84.72} & \best{70.74} & 77.78 & 84.15 & \best{45.00} & \best{89.11} & \best{73.36} & \best{79.04} \\
    Mask bottom $\rho$ fraction by $\igt$ (direction flip; drops lowest-shift positions) & \snd{76.23} & \snd{74.12} & 93.24 & 92.06 & \snd{83.91} & 68.21 & 80.20 & 83.61 & \snd{43.75} & \snd{88.22} & \snd{72.80} & \snd{78.35} \\
    Mask bottom $\rho$ fraction by $|\igt|$ (both tails; drops smallest absolute shifts) & 74.33 & 73.25 & \snd{93.85} & \best{92.49} & 83.48 & 66.53 & \best{81.03} & \snd{84.39} & 40.67 & 87.27 & 71.98 & 77.73 \\
    Random masking & 75.28 & 72.61 & \snd{93.85} & \snd{92.19} & 83.48 & 68.63 & 80.40 & \best{85.79} & 37.50 & \snd{88.22} & 72.11 & 77.80 \\
    Mask bottom $\rho$ fraction by student entropy & 73.64 & 73.20 & 93.43 & 92.06 & 83.08 & \snd{69.89} & \snd{81.01} & 83.06 & 38.12 & 87.78 & 71.97 & 77.53 \\
    Mask bottom $\rho$ fraction by teacher entropy & 74.07 & 72.95 & 93.30 & 91.79 & 83.03 & 67.79 & 78.18 & 82.51 & 43.12 & \snd{88.22} & 71.97 & 77.50 \\
    \bottomrule
  \end{tabular}%
  }
\end{table}

At 4B, top-$\igt$ masking obtains a total average of 79.04, compared with 78.35 for the direction-flipped selector and at most 77.80 for the $|\igt|$, random, student-entropy, and teacher-entropy controls. As at 30B, these results support selecting positions by signed entropy shift, with lower-shift selection outperforming the tested alternatives.

\section{Implementation Details}
\label{app:implementation}

\paragraph{Training framework.}
All runs use the \texttt{verl}~\citep{sheng2024hybridflow}\footnote{Apache-2.0 license.} on-policy RL framework with FSDP~\citep{zhao2023fsdp} for parameter/optimizer sharding and vLLM~\citep{kwon2023vllm}\footnote{Apache-2.0 license.} for rollouts.

\paragraph{Optimizer and schedule.}
We use AdamW~\citep{loshchilov2019adamw} with a constant learning-rate schedule (no warmup).

\paragraph{Common training hyperparameters.}
The following are shared across all six training runs (Qwen3-4B-Instruct and Qwen3-30B-A3B-Instruct, each with \drgrpo{}, naive \sdpo{}, and \sdpo{}+mask):
train batch size $256$ (mini-batch and PPO mini-batch both equal to $256$);
$3$ training epochs;
same order of training samples across runs;
max prompt length $4096$, max response length $8192$;
per-epoch A/B response-order swap enabled.

\paragraph{Per-method hyperparameters.}
\drgrpo{} uses learning rate $1\mathrm{e}{-}6$ (4B) / $2\mathrm{e}{-}6$ (30B) and $8$ rollouts per prompt.
Naive \sdpo{} and \sdpo{}+mask both use learning rate $5\mathrm{e}{-}6$ (4B) / $1\mathrm{e}{-}5$ (30B), $4$ rollouts per prompt, and $k=100$. To reduce computational cost, we approximate the full-vocabulary reverse KL using the teacher's top-$k$ tokens and a tail-remainder bucket~\citep{huebotter2026sdpo}. Student and teacher entropies used to compute $\igt$ are calculated from their full-vocabulary distributions. The teacher parameters are maintained as an exponential moving average (EMA) of the student parameters, with an update rate of $0.01$, to stabilize training.

\paragraph{Inference / evaluation decoding.}
At evaluation we sample one rollout per prompt with temperature $0.7$, top-$p$ $0.8$, and top-$k$ $20$, following the official Qwen3-Instruct model cards. We use the prompt template in Appendix~\ref{app:prompt}.

\section{Prompt Templates}
\label{app:prompt}

\paragraph{Input format.}
We send a single user message of the form shown in the templates below; no system prompt is used. The judge expects each turn of the dialog and each candidate response to be wrapped with \texttt{<user>...</user>} and \texttt{<assistant>...</assistant>} tags.

The placeholder \texttt{\{context\}} is the user-side input: for a single-turn query it is just one \texttt{<user>...</user>} block; for a multi-turn conversation it is the full alternating dialog (which must alternate user, assistant, user, $\ldots$ and end on a \texttt{<user>} turn). The placeholders \texttt{\{response\_a\}} and \texttt{\{response\_b\}} are the two candidate replies, each wrapped in a single \texttt{<assistant>} block. The teacher prompt inserts the language feedback at \texttt{\{feedback\}}. This field is omitted from the student and evaluation prompts.

\paragraph{Rationale language-feedback template.}
\begin{quote}\small\ttfamily
You are an impartial judge tasked with determining which of two assistant responses is better for the given context.\\[2pt]
Below is a context (a user query or a conversation between the user and an assistant) and two assistant responses to that context.\\[2pt]
{}[Start of Context]\\
\{context\}\\
{}[End of Context]\\[2pt]
{}[Start of Assistant A's Response]\\
\{response\_a\}\\
{}[End of Assistant A's Response]\\[2pt]
{}[Start of Assistant B's Response]\\
\{response\_b\}\\
{}[End of Assistant B's Response]\\[2pt]
\\
\{feedback\}
\\
\\[2pt]
Identify the quality dimensions that matter most for this specific task, then evaluate and compare the two assistant responses step by step across those dimensions. When correctness matters, solve the problem yourself and check each response for any errors. After your analysis, determine which response is better overall and provide your final verdict (A or B only) in <verdict>...</verdict>.
\end{quote}

\paragraph{Rubric language-feedback template.}
The rubric-feedback experiments use the same prompt, except for the final instruction paragraph:
\begin{quote}\small\ttfamily
You are an impartial judge tasked with determining which of two assistant responses is better for the given context.\\[2pt]
Below is a context (a user query or a conversation between the user and an assistant) and two assistant responses to that context.\\[2pt]
{}[Start of Context]\\
\{context\}\\
{}[End of Context]\\[2pt]
{}[Start of Assistant A's Response]\\
\{response\_a\}\\
{}[End of Assistant A's Response]\\[2pt]
{}[Start of Assistant B's Response]\\
\{response\_b\}\\
{}[End of Assistant B's Response]\\[2pt]
\\
\{feedback\}
\\
\\[2pt]
Identify the rubric that matters most for this specific task: the hard requirements the response must satisfy, ranked by importance, and the discriminative criteria that most decisively separate a better response from a worse one, ranked most decisive first, stating for each what makes a response better versus worse. Then evaluate and compare the two assistant responses step by step against that rubric. When correctness matters, solve the problem yourself and check each response for any errors. After your analysis, determine which response is better overall and provide your final verdict (A or B only) in <verdict>...</verdict>.
\end{quote}

\paragraph{Single-turn example.}
\begin{quote}\small\ttfamily
{}[Start of Context]\\
<user>\\
What is the capital of France?\\
</user>\\
{}[End of Context]\\[2pt]
{}[Start of Assistant A's Response]\\
<assistant>\\
The capital of France is Paris.\\
</assistant>\\
{}[End of Assistant A's Response]\\[2pt]
{}[Start of Assistant B's Response]\\
<assistant>\\
Lyon.\\
</assistant>\\
{}[End of Assistant B's Response]
\end{quote}

\paragraph{Multi-turn example.}
\begin{quote}\small\ttfamily
{}[Start of Context]\\
<user>\\
I'm planning a 3-day trip to Tokyo next month. Any recommendations?\\
</user>\\[2pt]
<assistant>\\
Sure, what kind of activities are you interested in (food, history, nightlife, shopping)?\\
</assistant>\\[2pt]
<user>\\
Mostly food and history.\\
</user>\\
{}[End of Context]\\[2pt]
{}[Start of Assistant A's Response]\\
<assistant>\\
Day 1: Tsukiji outer market for breakfast $\ldots$\\
</assistant>\\
{}[End of Assistant A's Response]\\[2pt]
{}[Start of Assistant B's Response]\\
<assistant>\\
Just go to Shibuya and figure it out when you get there.\\
</assistant>\\
{}[End of Assistant B's Response]
\end{quote}